\documentclass{article}
\usepackage[accepted]{icml2019}

\usepackage{microtype}      %
\usepackage{graphicx}
\usepackage{booktabs} 		%

\usepackage[utf8]{inputenc} %
\usepackage[T1]{fontenc}    %
\usepackage{hyperref}       %
\usepackage{url}            %
\usepackage{nicefrac}       %

\usepackage{multirow}
\usepackage{amsmath}
\usepackage{amsfonts}
\usepackage{amssymb}
\usepackage{tikz}
\usepackage{pgfplots}
\usetikzlibrary{shapes,arrows}
\usetikzlibrary{matrix,chains,shapes.geometric,positioning,decorations.pathreplacing,arrows}
\usetikzlibrary{arrows,calc,decorations.markings,math,arrows.meta}
\usepackage{subcaption} 
\usepackage{placeins}
\usepackage{marginnote}

\usepackage{mlmacros}		%

\variables{x,z,u,s,A,B}
\variables[mean,std]{\mu,\sigma}
\varmacros{E}{\mathcal{E}}

\probdists{p,q}
\probdists[policy]{\pi}
\MkProbDist{P}{\mathbb{P}}

\icmltitlerunning{Switching Linear Dynamics for Variational Bayes Filtering}

\begin{document}

\twocolumn[
\icmltitle{Switching Linear Dynamics for Variational Bayes Filtering}

\icmlsetsymbol{equal}{*}

\begin{icmlauthorlist}
\icmlauthor{Philip Becker-Ehmck}{vw,tud}
\icmlauthor{Jan Peters}{tud}
\icmlauthor{Patrick van der Smagt}{vw}
\end{icmlauthorlist}

\icmlaffiliation{vw}{Machine Learning Research Lab, Volkswagen Group, Munich, Germany}
\icmlaffiliation{tud}{Department of Computer Science, Technische Universität Darmstadt, Darmstadt, Germany}

\icmlcorrespondingauthor{Philip Becker-Ehmck}{philip.becker-ehmck@volkswagen.de}

\icmlkeywords{Machine Learning, ICML}

\vskip 0.3in
]

\printAffiliationsAndNotice{}  %

\begin{abstract}
System identification of complex and nonlinear systems is a central problem for model predictive control and model-based reinforcement learning. 
Despite their complexity, such systems can often be approximated well by a set of linear dynamical systems if broken into appropriate subsequences. 
This mechanism not only helps us find good approximations of dynamics, but also gives us deeper insight into the underlying system. 
Leveraging Bayesian inference, Variational Autoencoders and Concrete relaxations, we show how to learn a richer and more meaningful state space, e.g.\ encoding joint constraints and collisions with walls in a maze, from partial and high-dimensional observations.
This representation translates into a gain of accuracy of learned dynamics showcased on various simulated tasks.
\end{abstract}

\section{Introduction}
\label{introduction}
Learning dynamics from raw data (also known as system identification) is a key component of model predictive control and model-based reinforcement learning. 
Problematically, environments of interest often give rise to very complex and highly nonlinear dynamics which are seemingly difficult to approximate. 
However, switching linear dynamical systems (SLDS) approaches claim that those environments can often be broken down into simpler units made up of areas of equal and linear dynamics \citep{ackerson1970state, chang1978state}.

Not only have those approaches demonstrated good predictive performance in various settings, which often is the sole goal of learning a system's dynamics, they also encode useful information into so called \emph{switching variables} which determine the dynamics of the next transition.
For example, when looking at the movement of an arm, one is intuitively aware of certain restrictions of possible movements, e.g., constraints to the movement due to joint constraints or obstacles.
The knowledge is present without the need to simulate; it is explicit. 
Exactly this kind of information will be encoded when successfully inferring switching variables.
Our goal in this work will therefore entail the search for richer representations in the form of latent state space models which encode knowledge about the underlying system dynamics. 
In turn, we expect this to improve the accuracy of our simulation as well. 
Such a representation alone could then be used in a reinforcement learning approach that possibly only takes advantage of the learned latent features but not necessarily its learned dynamics.

To learn richer representations, we identify one common problem with prevalent recurrent Variational Autoencoder models \citep{chung2015recurrent, fraccaro2016sequential, karl2017deep, Krishnan2015}: the non-probabilistic treatment of the transition dynamics often modelled by a powerful nonlinear function approximator.
From the history of the Autoencoder to the Variational Autoencoder, we know that in order to detect robust features in an unsupervised manner, probabilistic treatment of the latent space is paramount \citep{dai2017hidden,kingma2014semi}.
As our starting point, we will build on previously proposed approaches by \citet{krishnan2017structured} and \citet{karl2017deep}.
The latter already made use of locally linear dynamics, but only in a deterministic fashion. 
We extend their approaches by a stochastic SLDS model with structured inference and show that such treatment is vital for learning richer representations and simulation accuracy.

\section{Background}\label{background}

We consider discretized time-series data consisting of continuous observations $x_t \in \mcX \subset \RR^{n_x}$ and control inputs $u_t \in \mcU \subset \RR^{n_u}$ that we would like to model by corresponding latent states $z_t \in \mcZ \subset \RR^{n_z}$. 
We will denote sequences of variables by $x\Ts = (x_1, x_2, ..., x_T)$. 

\begin{figure*}[t]
  \centering 
  \begin{subfigure}[b]{0.49\textwidth}
  \centering
   \resizebox {.7\textwidth} {!} {
    \begin{tikzpicture}[
>={Stealth[scale=2.0]},
var/.style={
draw,
fill=white,
circle, 
minimum width={width("$z_{t+1}$")+2pt},
minimum height={height("$z_{t+1}$")+2pt},
font=\large},
input/.style={
draw,
fill=lightgray,
circle, 
minimum width={width("$z_{t+1}$")+2pt},
minimum height={height("$z_{t+1}$")+2pt},
font=\large},
det/.style={
draw,
fill=white,
diamond, 
font=\large},
network/.style={
draw,
fill=Aquamarine,
rectangle, 
minimum width={width("$z_{t+1}$")+4pt},
minimum height={height("$z_{t+1}$")+4pt},
font=\large}]

\def\tzero{3}
\def\tone{6}
\def\ttwo{9}

\node[var, draw=black!100] at (\tzero, 6) (s_1) {$s_{1}$};
\node[var, draw=black!100] at (\tone, 6) (s_2) {$s_{2}$};
\node[var, draw=black!100] at (\ttwo, 6) (s_3) {$s_{3}$};

\node[var, draw=black!100] at (\tzero, 4) (z_1) {$z_1$};
\node[var, draw=black!100] at (\tone, 4) (z_2) {$z_2$};
\node[var, draw=black!100] at (\ttwo, 4) (z_3) {$z_3$};

\node[input, draw=black!100] at (\tzero +1.5, 5) (u_1) {$u_1$};
\node[input, draw=black!100] at (\tone + 1.5, 5) (u_2) {$u_2$};

\node[input, draw=black!100] at (\tzero, 2) (x_1) {$x_1$};
\node[input, draw=black!100] at (\tone, 2) (x_2) {$x_2$};
\node[input, draw=black!100] at (\ttwo, 2) (x_3) {$x_3$};

\node at (6, 2.5) (F) {};
\coordinate (f_1) at ([xshift=1.cm]F.east);

\draw (z_1) edge[->] (x_1);
\draw (z_1) edge[->] (z_2);
\draw[->,out=60,in=190] (z_1) to (s_2);
\draw (z_2) edge[->] (x_2);
\draw (z_2) edge[->] (z_3);
\draw[->,out=60,in=190] (z_2) to (s_3);
\draw (z_3) edge[->] (x_3);

\draw (u_1) edge[->] (z_2);
\draw (u_2) edge[->] (z_3);
\draw (u_1) edge[->] (s_2);
\draw (u_2) edge[->] (s_3);

\draw (s_1) edge[->] (s_2);
\draw (s_1) edge[->] (z_1);
\draw (s_2) edge[->] (s_3);
\draw (s_2) edge[->] (z_2);
\draw (s_3) edge[->] (z_3);

\draw[->,in=60,out=300] (s_1) to (x_1);
\draw[->,in=60,out=300] (s_2) to (x_2);
\draw[->,in=60,out=300] (s_3) to (x_3);

\end{tikzpicture}
   }
    \caption{SLDS graphical model.}
    \label{fig:slds-generative}
  \end{subfigure}
  \begin{subfigure}[b]{0.49\textwidth}
  \centering
 	\resizebox {.7\textwidth} {!} {
    \begin{tikzpicture}[
>={Stealth[scale=2.0]},
var/.style={
draw,
fill=white,
circle, 
minimum width={width("$z_{t+1}$")+2pt},
minimum height={height("$z_{t+1}$")+2pt},
font=\large},
input/.style={
draw,
fill=lightgray,
circle, 
minimum width={width("$z_{t+1}$")+2pt},
minimum height={height("$z_{t+1}$")+2pt},
font=\large},
det/.style={
draw,
fill=white,
diamond, 
font=\large},
network/.style={
draw,
fill=Aquamarine,
rectangle, 
minimum width={width("$z_{t+1}$")+4pt},
minimum height={height("$z_{t+1}$")+4pt},
font=\large}]

\def\tzero{3}
\def\tone{6}
\def\ttwo{9}

\node[var, draw=black!100] at (\tzero, 6) (w) {$h$};
\node[var, draw=black!100] at (\tone, 6) (s_2) {$s_2$};
\node[var, draw=black!100] at (\ttwo, 6) (s_3) {$s_3$};

\node[det, draw=black!100] at (\tzero, 4) (z_1) {$z_1$};
\node[var, draw=black!100] at (\tone, 4) (z_2) {$z_2$};
\node[var, draw=black!100] at (\ttwo, 4) (z_3) {$z_3$};

\node[input, draw=black!100] at (\tzero +1.5, 5) (u_1) {$u_1$};
\node[input, draw=black!100] at (\tone + 1.5, 5) (u_2) {$u_2$};

\node[input, draw=black!100] at (\tzero, 2) (x_1) {$x_1$};
\node[input, draw=black!100] at (\tone, 2) (x_2) {$x_2$};
\node[input, draw=black!100] at (\ttwo, 2) (x_3) {$x_3$};

\node at (6, 2.5) (F) {};
\coordinate (f_1) at ([xshift=1.cm]F.east);

\draw (z_1) edge[->] (x_1);
\draw (z_2) edge[->] (x_2);
\draw (z_3) edge[->] (x_3);

\draw (z_1) edge[->] (z_2);
\draw[->,out=60,in=190] (z_1) to (s_2);
\draw (z_2) edge[->] (z_3);
\draw[->,out=60,in=190] (z_2) to (s_3);

\draw (u_1) edge[->] (z_2);
\draw (u_2) edge[->] (z_3);
\draw (u_1) edge[->] (s_2);
\draw (u_2) edge[->] (s_3);

\draw (w) edge[->] (z_1);
\draw (s_2) edge[->] (s_3);
\draw (s_2) edge[->] (z_2);
\draw (s_3) edge[->] (z_3);

\end{tikzpicture}
    }
    \caption{Our generative model.}
    \label{fig:generative}
  \end{subfigure}
  \caption{(a) $s_t$ denote discrete switch variables, $z_t$ are continuous latent variables, $x_t$ are continuous observed variables, $u_t$ are (optional) continuous control inputs. (b) By introducing a special latent variable $h$ used for initial state inference, we want to make explicit that the first step is treated differently from the rest of the sequence.}
\end{figure*}
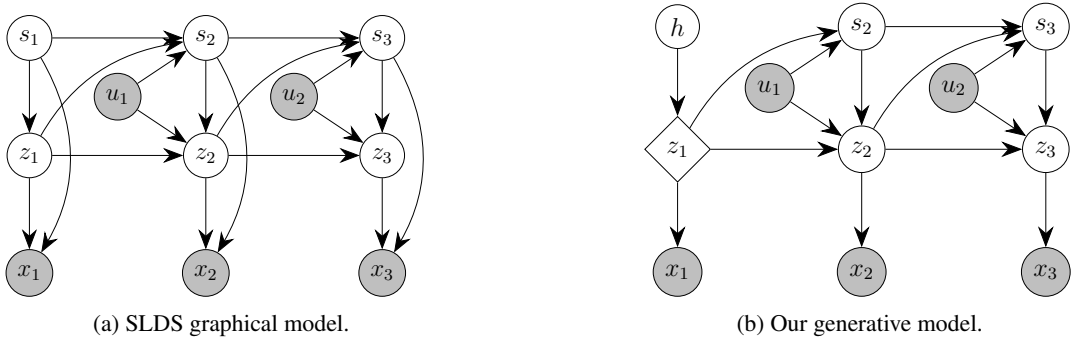

\subsection{Switching Linear Dynamical Systems}
A Switching Linear Dynamical Systems (SLDS) model enables us to model nonlinear time series data by splitting it into subsequences governed by linear dynamics. 
At each time $t=1,2,...,T$, a discrete switch variable $s_t \in 1,...,M$ chooses of a set of $M$ LDSs a system which is to be used to transform the continuous latent state $z\tm$ to the next time step \citep{barberBRML2012}:
\begin{equation}
\label{eq:slds-transition}
\begin{aligned}
z_t &= A(s_t) z\tm + B(s_t) u\tm + \epsilon(s_t) \\
x_t &= H(s_t)z_t + \eta(s_t) \\
\text{with } \quad &\eta(s_t) \sim \gauss{0,R(s_t)},\: \epsilon(s_t) \sim \gauss{0,Q(s_t)}
\end{aligned}
\end{equation}
Here $A \in \RR^{n_z \times n_z}$ is the state matrix, $B \in \RR^{n_z \times n_u}$ is the control matrix, $\epsilon$ is the transition noise with covariance matrix $Q$, $\eta$ is the emission/sensor noise with covariance matrix $R$ and the observation matrix $H \in \RR^{n_x \times n_z}$ defines a linear mapping from latent to observation space.
These equations imply the following joint distribution:
\begin{equation}
\label{eq:slds-joint}
\begin{aligned}
&\p{x\Ts, z\Ts, s\Ts}{u\Ts} = \prod_{t=1}^T \p{x_t}{z_t,s_t}
\\ & \qquad \p{z_t}{z\tm,u\tm,s_t} \, \p{s_t}{z\tm,u\tm,s\tm}
\end{aligned}
\end{equation} 
with $\p{s_1}{z_0, u_0, s_0} = p(s_1)$ and $\p{z_1}{z_0, u_0, s_1} = \p{z_1}{s_1}$ being initial state distributions. The discrete switching variables are usually assumed to evolve according to Markovian dynamics, i.e.\ $\mathrm{Pr}(s_t| s\tm = k) = \pi_k$, which optionally may be conditioned on the continuous state $z\tm$. The corresponding graphical model is shown in figure~\ref{fig:slds-generative}.

\subsection{Stochastic Gradient Variational Bayes}
\begin{equation}
\p{x} = \int \p{x,z} \dint z = \int \p{x}{z}\p{z} \dint z 
\label{eq:vae-graphical-model}
\end{equation} 
Given the simple graphical model in equation~(\ref{eq:vae-graphical-model}), \citet{kingma2014auto} and \citet{Rezende2014} introduced the \textit{Variational Autoencoder} (VAE) which overcomes the intractability of posterior inference of $\p{z}{x}$ by maximizing the evidence lower bound (ELBO) of the model log-likelihood:
\begin{equation}
\label{eq:vae-elbo}
\begin{aligned}
&\log \p{x} \geq \loss[\mathrm{ELBO}]{x; \theta, \phi}\\
&= \expc[\q[\phi]{z}{x}]{\ln \p[\theta]{x}{z}} - \kl{\q[\phi]{z}{x}}{\p{z}}.
\end{aligned}
\end{equation}
Their main innovation was to approximate the intractable posterior $\p{z}{x}$ by a \textit{recognition network} $q_\phi(z|x)$ from which they can sample via the \textit{reparameterization trick} to allow for stochastic backpropagation through both the recognition and generative model at once. 
Assuming that the latent state is normally distributed, a simple transformation allows us to obtain a Monte Carlo gradient estimate of $\mathbb{E}_{q_\phi(z|x)} \left[ \ln p_\theta(x|z) \right]$ w.r.t.\ $\phi$. 
Given that $z \sim \mathcal{N}(\mu, \sigma^2)$, we can generate samples by drawing from an auxiliary variable $\epsilon \sim \mathcal{N}(0,1)$ and applying the deterministic and differentiable transformation $z = \mu + \sigma \epsilon$.

\subsection{The Concrete Distribution}
One simple and efficient way to obtain samples $d$ from a $k$-dimensional categorical distribution with class probabilities $\alpha$ is the Gumbel-Max trick:
\begin{equation}
\begin{aligned}
d = \text{one\_hot} \left( \text{argmax} [g_i + \log \alpha_i ] \right)\\ \text{with } \: g_1, \ldots, g_k \sim \text{Gumbel}(0,1).
\end{aligned}
\end{equation}
However, since the derivative of the argmax is 0 everywhere except at the boundary of state changes, where it is undefined, we cannot learn a parameterization by backpropagation. 
The Gumbel-Softmax trick approximates the argmax by a softmax which gives us a probability vector \citep{Jang2017, Maddison2016}. We can then draw samples via 
\begin{equation}
d_k = \frac{\exp ((\log \alpha_k + g_k) / \lambda)}{\sum_{i=1}^n \exp ((\log \alpha_i + g_i) / \lambda)}.
\end{equation}
This softmax computation approaches the discrete argmax as temperature $\lambda \rightarrow 0$ while for $\lambda \rightarrow \infty$ it approaches a uniform distribution. In the same vein, the bias of the estimator decreases with decreasing $\lambda$ while its variance increases.

\section{Related Work}

Our model can be viewed as a Deep Kalman Filter \citep{Krishnan2015} with structured inference \citep{krishnan2017structured}. 
In our case, structured inference entails another stochastic variable model with parameter sharing inspired by \citet{karl2017unsupervised} and \citet{karl2017deep} which pointed out the importance of backpropagating the reconstruction error through the generative transition.
\Citet{marino2018general} adopts this idea by starting an iterative amortized inference scheme with the prior prediction. Iterative updates are then only conditioned on the gradient of the loss, allowing the observation only to adjust, but not completely overrule, the prior prediction.
We are different to a number of stochastic sequential models like \citet{bayer2014learning, chung2015recurrent, goyal2017z, shabanian_variational_2017} by directly transitioning the stochastic latent variable over time instead of having an RNN augmented by stochastic inputs.
\citet{fraccaro2016sequential} proposes a transition over both a deterministic and a stochastic latent state sequence, wanting to combine the best of both worlds.

Previous models \citep{fraccaro2017disentangled, karl2017deep, watter2015embed} have already combined locally linear models with recurrent Variational Autoencoders, however they provide a weaker structural incentive for learning latent variables determining the transition function.
\Citet{steenkiste2018relational} approach a similar multi bouncing ball problem (see section~\ref{sec:bouncing-ball}) by first distributing the representation of different balls into their own entities without supervision and then structurally hardwiring a transition function with interactions based on an attention mechanism.

Recurrent switching linear dynamical systems \citep{Linderman2016} uses message passing for approximate inference, but has restricted itself to low-dimensional observations and a multi-stage training process.
\citet{nassar2018tree} builds on this work and suggests a tree structure for enforcing a locality prior on switching variables where subtrees share similar dynamics.
\citet{johnson2016composing} propose a similar model to ours but combine message passing for discrete switching variables with a neural network encoder for observations learned by stochastic backpropagation.

One feature an SLDS model may learn are interactions which have recently been approached by employing Graph Neural Networks \citep{battaglia2016interaction, kipf2018neural}. These methods are similar in that they predict edges which encode interactions between components of the state space (nodes).
Tackling the problem of propagating state uncertainty over time, various combinations of neural networks for inference and Gaussian processes for transition dynamics have been proposed \citep{doerr2018probabilistic, eleftheriadis2017identification}.
However, these models have not been demonstrated to work with high-dimensional observation spaces like images.

\section{Proposed Approach}
We propose learning an SLDS model through a recurrent Variational Autoencoder framework which approximates switching variables by a Concrete distribution \citep{Jang2017, Maddison2016}.
This leads to a model that can be optimized entirely by stochastic backpropagation through time.
For inference, we propose a time-factorized approach with a specific computational structure, reusing the generative model. 
This allows us to learn good transition dynamics.
Our generative model is shown in figure~\ref{fig:generative} an our inference model in figure~\ref{fig:inference}.

\subsection{Generative Model}

Our generative model for a single $x_t$ is described by 
\begin{gather}
\label{eq:generative-model}
\begin{aligned}
\p{x_t} = \int_{s_{\leq t}} \int_{z_{\leq t}} &\p{x_t}{z_t} \p{z_t}{z\tm, s_t, u\tm} \\ &\p{s_t}{s\tm, z\tm, u\tm} \p{z\tm, s\tm}
\end{aligned}
\raisetag{36pt}
\end{gather}
which is close to the original SLDS model (see figure~\ref{fig:slds-generative}).
Latent states $z_t$ are continuous and represent the state of the system while states $s_t$ are switching variables determining the transition. 
In order to use end-to-end backpropagation, we approximate the discrete switching variables by a continuous relaxation, namely the Concrete distribution.
Differently to the original model, we do not condition the likelihood of the current observation $\p[\theta]{x_t}{z_t}$ directly on the switching variables.
This limits the influence of the switching variables to choosing a proper transition dynamic for the continuous latent space.
The likelihood model is parameterized by a neural network with, depending on the data, either a Gaussian or a Bernoulli distribution as output.

As outlined in~\eqref{eq:generative-model}, we need to learn separate transition functions for the continuous states $z_t$ and for the discrete states $s_t$.
For the continuous state transition $\p{z_t}{z\tm, s_t, u\tm}$ we follow \eqref{eq:slds-transition} and maintain a set of $M$ base matrices $\lbrace \left( A^{(i)},B^{(i)},Q^{(i)} \right)| \: \forall i,\; 0 < i < M \rbrace$ as our linear dynamical systems to choose from: 
\begin{equation}
\begin{gathered}
\p[\theta]{z_t}{z\tm, s_t, u\tm} = \gauss{\mu, \sigma^2} \\
\text{where } \; \mu = A_\theta(s_t) z\tm + B_\theta(s_t) u_t, \: \sigma^2 = Q_\theta(s_t).\\
\end{gathered}
\end{equation}
For the transition on discrete latent states $\p{s_t}{s\tm, z\tm, u\tm}$, we conventionally require a Markov transition matrix.
However, since we approximate our discrete switching variables by a continuous relaxation, we can parameterize this transition by a neural network:
\begin{equation}
\begin{gathered}
\p[\theta]{s_t}{s\tm, z\tm, u\tm} = \text{Concrete}(\alpha, \lambda_{\mathrm{prior}}) \\
\text{where } \; \alpha = g_\theta(z\tm,s\tm,u\tm).
\end{gathered}
\end{equation}
Finally, the question arises how we determine our transition matrices $A, B$ and $Q$ since our Concrete samples $s_t$ are now probability vectors and not one-hot vectors anymore.
We could execute the forward pass by choosing the linear system corresponding to the highest value in the sample (hard forward pass) and only use the relaxation for our backward pass. 
This, however, means that we would follow a biased gradient.
Alternatively, we can use the relaxed version for our forward pass and aggregate the linear systems based on their corresponding weighting:
\begin{equation}
\begin{gathered}
\label{eq:linear-combination}
A_t(s_t) = \sum_{i=1}^M{s_t^{(i)} A^{(i)}}, \: B_t(s_t) = \sum_{i=1}^M{s_t^{(i)} B^{(i)}},\\
Q_t(s_t) = \sum_{i=1}^M{s_t^{(i)} Q^{(i)}}.
\end{gathered}
\end{equation}
Here, we lose the discrete switching of linear systems, but maintain a valid lower bound. 
We note that the hard forward pass has led to worse results and focus on the soft forward pass for this paper.

\subsection{Inference}
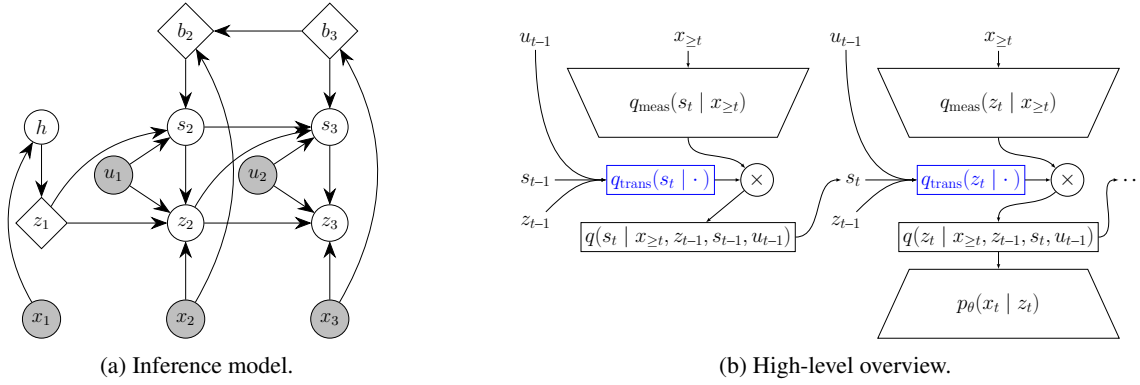
\begin{figure*}[t]
  \centering 
  \begin{subfigure}[b]{0.39\textwidth}
  \centering
   \resizebox {.8\textwidth} {!} {
    \begin{tikzpicture}[
>={Stealth[scale=2.0]},
var/.style={
draw,
fill=white,
circle, 
minimum width={width("$z_{t+1}$")+2pt},
minimum height={height("$z_{t+1}$")+2pt},
font=\large},
input/.style={
draw,
fill=lightgray,
circle, 
minimum width={width("$z_{t+1}$")+2pt},
minimum height={height("$z_{t+1}$")+2pt},
font=\large},
det/.style={
draw,
fill=white,
diamond, 
font=\large},
network/.style={
draw,
fill=Aquamarine,
rectangle, 
minimum width={width("$z_{t+1}$")+4pt},
minimum height={height("$z_{t+1}$")+4pt},
font=\large}]

\def\tzero{3}
\def\tone{6}
\def\ttwo{9}
\def\tend{13}

\node[det, draw=black!100] at (\tone, 8) (b_2) {$b_2$};
\node[det, draw=black!100] at (\ttwo, 8) (b_3) {$b_3$};

\node[var, draw=black!100] at (\tzero, 6) (w) {$h$};
\node[var, draw=black!100] at (\tone, 6) (s_2) {$s_2$};
\node[var, draw=black!100] at (\ttwo, 6) (s_3) {$s_3$};

\node[det, draw=black!100] at (\tzero, 4) (z_1) {$z_1$};
\node[var, draw=black!100] at (\tone, 4) (z_2) {$z_2$};
\node[var, draw=black!100] at (\ttwo, 4) (z_3) {$z_3$};

\node[input, draw=black!100] at (\tzero +1.5, 5) (u_1) {$u_1$};
\node[input, draw=black!100] at (\tone + 1.5, 5) (u_2) {$u_2$};

\node[input, draw=black!100] at (\tzero, 2) (x_1) {$x_1$};
\node[input, draw=black!100] at (\tone, 2) (x_2) {$x_2$};
\node[input, draw=black!100] at (\ttwo, 2) (x_3) {$x_3$};

\node at (6, 2.5) (F) {};
\coordinate (f_1) at ([xshift=1.cm]F.east);

\draw (x_1) edge[->,in=240,out=120] (w);
\draw (x_2) edge[->] (z_2);
\draw (x_3) edge[->] (z_3);

\draw (z_1) edge[->] (z_2);
\draw[->,out=60,in=190] (z_1) to (s_2);
\draw (z_2) edge[->] (z_3);
\draw[->,out=60,in=190] (z_2) to (s_3);

\draw (u_1) edge[->] (z_2);
\draw (u_2) edge[->] (z_3);
\draw (u_1) edge[->] (s_2);
\draw (u_2) edge[->] (s_3);

\draw (w) edge[->] (z_1);
\draw (s_2) edge[->] (s_3);
\draw (s_2) edge[->] (z_2);
\draw (s_3) edge[->] (z_3);

\draw[->,in=300,out=60] (x_2) to (b_2);
\draw[->,in=300,out=60] (x_3) to (b_3);

\draw (b_2) edge[->] (s_2);
\draw (b_3) edge[->] (s_3);

\draw (b_3) edge[->] (b_2);

\end{tikzpicture}
    }
    \caption{Inference model.}
    \label{fig:inference}
  \end{subfigure}
  \begin{subfigure}[b]{0.59\textwidth}
  \centering
  \resizebox {.85\textwidth} {!} {
    \begin{tikzpicture} %
\tikzset{
  font={\fontsize{18pt}{20}\selectfont}}
\tikzset{trapezium stretches=true}
\tikzset{>=latex}

\node [trapezium, trapezium angle=100, minimum width=7cm, minimum height=2cm, draw] (rec) at (0,0) {};
\node at (rec.center) {$\q[\mathrm{meas}]{z_t}{x_{\geq t}}$};

\node [trapezium, trapezium angle=100, minimum width=7cm, minimum height=2cm, draw] (srec) at (-9,0) {};
\node at (srec.center) {$\q[\mathrm{meas}]{s_t}{x_{\geq t}}$};

\node [below=.75cm of rec, circle, draw, xshift=2cm] (mult) {$\times$};
\node[rectangle, draw, color=blue, left=.75cm of mult, align=center] (tra) {$\q[\mathrm{trans}]{z_t}{\cdot\:} $};
\node[rectangle, draw, minimum width=5.25cm, below=.75cm of mult, xshift=-2cm] (filter) { $\q{z_t}{x_{\geq t},z\tm, s_t, u\tm}$};

\node [below=.75cm of srec, circle, draw, xshift=2cm] (switchingmult) {$\times$};
\node[rectangle, draw, color=blue, left=.75cm of switchingmult, align=center] (switchingtra) {$\q[\mathrm{trans}]{s_t}{\cdot\:} $};

\node [trapezium, trapezium angle=80, minimum width=7cm, minimum height=2cm, draw, below=.5cm of filter] (gen) {};
\node at (gen.center) {$\p[\theta]{x_t}{z_t}$};

\node [below left=0.5cm and 1.5cm of switchingtra] (prev2) {$z\tm$};
\node [left=1.5cm of switchingtra] (switchingprev) {$s\tm$};
\node [below left=0.5cm and 1.5cm of tra] (prev) {$z\tm$};
\node [rectangle, draw, minimum width=5.25cm,below=.75cm of switchingmult, xshift=-2cm] (qs) {$\q{s_t}{x_{\geq t},z\tm, s\tm, u\tm}$};
\node [left=1.5cm of tra] (s) {$s_t$};
\node [above=.5cm of rec] (obs) {$x_{\geq t}$};
\node [above=.5cm of srec] (obs2) {$x_{\geq t}$};
\node at (prev |- obs) (con) {$u\tm$};
\node at (prev2 |- obs2) (con2) {$u\tm$};
\node [right=1cm of mult] (next) {\dots};

\path 
(rec.south) edge[->, out=270, in=135]  (mult)
(filter) edge[->]  (gen)
(prev) edge[->,out=45, in=180]  (tra)
(con) edge[->,out=270, in=180] (tra)
(tra) edge[->]  (mult)
(mult) edge[->, out=225, in=90]  (filter)
(filter) edge[->, out=0, in=180]  (next)
(obs) edge[->]  (rec);

\path
(srec.south) edge[->, out=270, in=135]  (switchingmult)
(obs2) edge[->] (srec)
(switchingprev) edge[->]  (switchingtra)
(prev2) edge[->,out=45, in=180]  (switchingtra)
(con2) edge[->,out=270, in=180] (switchingtra)
(switchingtra) edge[->] (switchingmult)
(switchingmult) edge[->] (qs)
(qs) edge[->, out=0, in=180] (s)
(s) edge[->] (tra);
\end{tikzpicture}
    }
    \caption{High-level overview.}
    \label{fig:fusion}
  \end{subfigure}
  \caption{(a) Depicts the inference model. $b_t$ is the hidden state of the backward RNN of $\q[\mathrm{meas}]{s_t}{x_{\geq t}, u_{\geq t}}$. Initial inference of $h$ may be conditioned on the entire sequence of observations, or just a subsequence. We've omitted the arrows for sake of clarity for the rest of the graph. (b) Shows schematically how we combine the transition with the inverse measurement model in the inference network. Transitions (in blue) are shared with the generative model.}
  \label{fig:inference-and-fusion}
\end{figure*}
As previously stated, the inference structure is critical for performance. 
In particular, we require the reconstruction loss gradient to flow through the generative transition model which is not naturally the case for these types of models.
Without a properly structured inference scheme, only the KL divergence would guide the generative transition model.
To achieve this, we formulate our inference scheme as a local optimization around the prior prediction where information from the observation only adjusts our prior prediction.
Generally, we split the inference model into two parts: 1) the generative transition model and 2) encoding the current (and optionally future) observations.
Both parts will independently predict a distribution over the next latent state which are then combined in a manner inspired by a Bayesian update.
In the following, we will go through the specific construction for both normal and Concrete distributions. 
The overall inference structure is depicted in figure~\ref{fig:fusion}. 

\subsubsection{Structured Inference of Gaussian Latent State}
Starting from the factorization of our true posterior, our approximate posterior takes the form $\q[\phi]{z_t}{z\tm, s_t, x_{\geq t}, u_{\geq \tminus 1}}$ where we notice that observations up to the last time step can be omitted as they are summarized by the last Markovian latent state $z\tm$.
As mentioned, the inference model  splits into two parts: 1) transition model $\q[\mathrm{trans}]{z_t}{z\tm, s_t, u\tm}$ and 2) inverse measurement model $\q[\mathrm{meas}]{z_t}{x_{\geq t}, u_{\geq t}}$ as previously proposed in \citet{karl2017unsupervised}. 
This split allows us to reuse our generative transition model in place of $\q[\mathrm{trans}]{z_t}{z\tm, s_t, u\tm}$. 
For practical reasons, we only share the computation of the transition mean $\mu_{\mathrm{trans}}$ but not the variance $\sigma_{\mathrm{trans}}^2$ between inference and generative model. 
Both parts, $\q[\mathrm{meas}]$ and $\q[\mathrm{trans}]$, will give us independent predictions about the new state $z_t$ which will be combined in a manner akin to a Bayesian update in a Kalman Filter:
\begin{gather}
\begin{aligned}
\q[\phi]{z_t}{z\tm, s_t, x_{\geq t}, u_{\geq \tminus 1}} &= \gauss{\mu_q, \sigma_q^2} \\
\propto \q[\mathrm{meas}]{z_t}{x_{\geq t}, u_{\geq t}} &\times \q[\mathrm{trans}]{z_t}{z\tm,s_t,u\tm} \\
\q[\mathrm{meas}]{z_t}{x_{\geq t}, u_{\geq t}} &= \gauss{\mu_{\mathrm{meas}}, \sigma_{\mathrm{meas}}^2} \\
\text{where } \; [\mu_{\mathrm{meas}},\sigma_{\mathrm{meas}}^2] &= h_\phi(x_{\geq t}, u_{\geq t})\\
\q[\mathrm{trans}]{z_t}{z\tm,s_t,u\tm} &= \gauss{\mu_{\mathrm{trans}}, \sigma_{\mathrm{trans}}^2} \\
\text{where } \: \mu_{\mathrm{trans}} = A_\theta(s_t) z\tm &+ B_\theta(s_t) u_t, \sigma_{\mathrm{trans}}^2 = Q_\phi(s_t)
\end{aligned}
\raisetag{50pt}
\end{gather}
The densities of $\q[\mathrm{meas}]$ and $\q[\mathrm{trans}]$ are multiplied resulting in another Gaussian density:
\begin{equation}
\begin{aligned}
\mu_q &= \frac{\mu_{\mathrm{trans}} \sigma_{\mathrm{meas}}^2 + \mu_{\mathrm{meas}} \sigma_{\mathrm{trans}}^2}{\sigma_{\mathrm{meas}}^2 + \sigma_{\mathrm{trans}}^2},\\
\sigma_q^2 &= \frac{\sigma_{\mathrm{meas}}^2 \sigma_{\mathrm{trans}}^2}{\sigma_{\mathrm{meas}}^2 + \sigma_{\mathrm{trans}}^2}.
\end{aligned}
\end{equation}

To enable online filtering, we can condition the inverse measurement model $\q[\mathrm{meas}]{z_t}{x_{\geq t}, u_{\geq t}}$ solely on the current observation $x_t$ instead of the entire remaining trajectory. 
We found empirically that despite being theoretically suboptimal, this can yield good results in many cases.
Naturally this is the case for actually Markovian settings which are plentiful in the physical world.
In this case, this methodology can be used for real-time state filtering in online feedback control.

For the initial state $z_1$ we do not have a conditional prior from the transition model as in the rest of the sequence.
Other methods \citep{Krishnan2015, fraccaro2016sequential} have used a standard normal prior, however this is not a good fit. 
We therefore decided that instead of predicting $z_1$ directly, we predict an auxiliary variable $h$ that is then mapped deterministically to a starting state $z_1$. 
A standard Gaussian prior is then applied to $h$:
\begin{equation}
\begin{aligned}
\q[\phi]{h}{x\tsub{1}{T},u\tsub{1}{T}} &= \gauss{h;\mu_w,\sigma_w^2} \\
\text{where} \quad [\mu_h,\sigma_h^2] &= i_\phi(x\tsub{1}{T},u\tsub{1}{T})\\
z_1 &= t_\phi(h)
\end{aligned}
\end{equation}
Alternatively, we could specify a more complex or learned prior for the initial state like the VampPrior \citep{tomczak2017vae}.
Empirically, we were unable to produce good results with these approaches.

\subsubsection{Inference of Switching Variables}

Following \citet{Maddison2016} and \citet{Jang2017}, we can reparameterize a discrete latent variable with the Gumbel-softmax trick. 
Again, we split our inference network $\q[\phi]{s_t}{s\tm, z\tm, x_{\geq t}, u_{\geq \tminus 1}}$ in an identical fashion into two components: 1) transition model $\q[\mathrm{trans}]{s_t}{s\tm, z\tm, u\tm}$ and 2) inverse measurement model $\q[\mathrm{meas}]{s_t}{x_{\geq t}, u_{\geq t}}$. 
The transition model is again shared with the generative model and is implemented via a neural network as we potentially require quick changes to chosen dynamics. The inverse measurement model is parametrized by a backward LSTM (or an MLP in the online filtering setting). 
For Concrete variables, we let each network predict the logits of a Concrete distribution and our inverse measurement model $\q[\phi]{s_t}{x_{\geq t}, u_{\geq t}}$ produces an additional vector $\gamma$, which determines the value of a gate deciding how the two predictions are to be weighted:
\begin{gather}
\begin{aligned}
&\q[\phi]{s_t}{s\tm, z\tm, x_{\geq t}, u_{\geq \tminus 1}} = \text{Concrete}(\alpha, \lambda_{\mathrm{posterior}}) \\ 
&\qquad\qquad\text{with} \quad \alpha = \gamma \alpha_{\mathrm{trans}} + (1-\gamma)\alpha_{\mathrm{meas}}  \\
&\q[\mathrm{meas}]{s_t}{x_{\geq t}, u_{\geq t}} = \text{Concrete}(\alpha_{\mathrm{meas}}, \lambda_{\mathrm{posterior}}) \\
&\qquad\qquad\text{where} \quad [\alpha_{\mathrm{meas}}, \gamma] = k_\phi(x_{\geq t}, u_{\geq t})\\
&\q[\mathrm{trans}]{s_t}{s\tm, z\tm,u\tm} = \text{Concrete}(\alpha_{\mathrm{trans}}, \lambda_{\mathrm{prior}})\\ 
&\qquad\qquad\text{where} \quad \alpha_{\mathrm{trans}} = g_\theta(z\tm,s\tm,u\tm)
\end{aligned}
\raisetag{52pt}
\end{gather}
The temperatures $\lambda_{\mathrm{posterior}}$ and $\lambda_{\mathrm{prior}}$ are set as a hyperparameter and can be set differently for the prior and approximate posterior.
The gating mechanism gives the model the option to balance between prior and approximate posterior. 
If the prior is good enough to explain the next observation, $\gamma$ will be pushed to 1 which will ignore the measurement and will minimize the KL between prior and posterior by only propagating the prior. 
If the prior is not sufficient, information from the inverse measurement model can flow by decreasing $\gamma$ and incurring a KL penalty.

\subsubsection{Reinterpretation as a Hierarchical Model}
Lastly, we could step away from the theory of SLDS and instead view our model simply as an hierarchical graphical model where some variables should explain the current observation while others should determine the transition over said variables. 
When viewed in this manner, it suggests to model the switching variables by any distribution, e.g.\ also as normally distributed with a specific decoder structure predicting some kind of transition parameters.
If this worked better, it would highlight still existing optimization problems of discrete random variables. 
Additionally, it could support our initial claim that stochastic treatment of the transition dynamics in general is important, irrespective of the specific implementation.
As such, it will act as an ablation study for our model.
Although any number of parameterizations are now viable, the one that we explore here is to let our decoder to predict mixing coefficients $\alpha$ for our transition matrices. We suggest just using a single linear layer:
\begin{equation}
\label{eq:softmax-alpha}
\begin{aligned}
\alpha = \text{softmax}(W s_t + b) \in \RR^M. \\
\end{aligned}
\end{equation}
In this scenario, our inference scheme for normally distributed switching variables is then identical to the one described in the previous section.

\begin{figure*}[t]
  \centering 
  \begin{subfigure}[b]{0.24\textwidth}
  \centering
   \includegraphics[scale=.2]{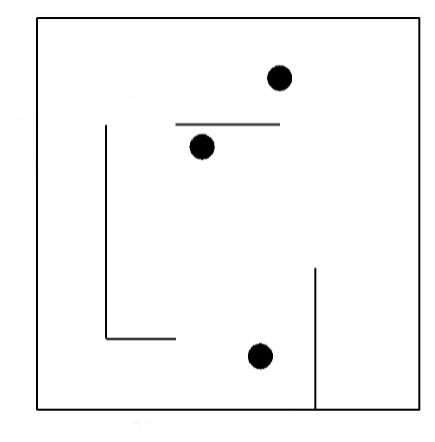}
    \caption{Multi agent maze environment.}
    \label{fig:maze-env}
  \end{subfigure}\hfill
  \begin{subfigure}[b]{0.24\textwidth}
  \centering
  	\includegraphics[scale=0.2]{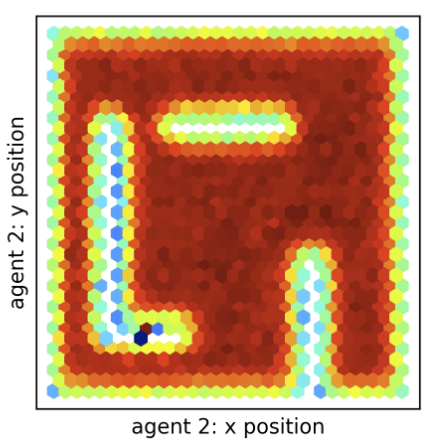}
    \caption{Variable encoding free space for agent 2.}
    \label{fig:maze-free}
  \end{subfigure}\hfill
  \begin{subfigure}[b]{0.24\textwidth}
  \centering
  	\includegraphics[scale=0.2]{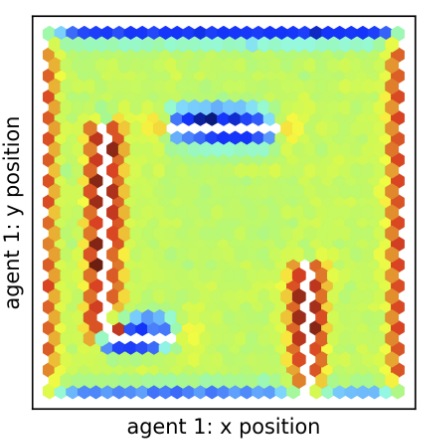}
    \caption{Variable encoding walls for agent 1.}
    \label{fig:maze-wall}
  \end{subfigure}\hfill
  \begin{subfigure}[b]{0.24\textwidth}
  \centering
  	\includegraphics[scale=0.205]{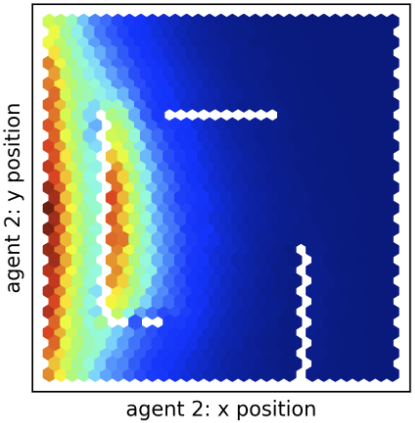}
    \caption{System activation for deterministic transition.}
    \label{fig:maze-deterministic}
  \end{subfigure}
  \caption{Figures (b) and (c) depict an agent's position coloured by the average value of a single latent variable $s$ marginalized over all controls and velocities. Figure (d) shows a typical activation map for a single transition system for deterministic treatment of transition dynamics. It does not generalize to the entire maze and stays fairly active near the wall.}
  \label{fig:maze}
\end{figure*}

\subsection{Training}
Our objective function is the commonly used evidence lower bound for our hierarchical model:
\begin{gather}
\begin{aligned}
&\loss[\theta,\phi]{x\Ts}{u\Ts} \geq\\
&\expc[\q[\phi]{z\Ts,s_{2:T}}{x\Ts}]{\log \p[\theta]{x\Ts}{z\Ts, s_{2:T}, u\Ts}} \\ 
&-\kl{\q[\phi]{z\Ts, s_{2:T}}{x\Ts, u\Ts}}{\p{z\Ts, s_{2:T}}{u\Ts}}.
\end{aligned}
\raisetag{30pt}
\end{gather}
As we chose to factorize over time, the loss for a single observation $x_t$ becomes:
\begin{gather}
\label{eq:loss}
\begin{aligned}
&\loss[\theta,\phi]{x_t}{u\Ts} = \expc[s_t]{\expc[z_t]{\log \p[\theta]{x_t}{z_t}}} \\ 
&- \expc[s_{t-1}, z_{t-1}]{\kl{\q[\phi]{s_t}{\cdot\:}}{\p[\theta]{s_t}{s\tm,z\tm,u\tm}}}\\
&- \expc[z_{t-1}]{\expc[s_t]{\kl{\q[\phi]{z_t}{\cdot\:}}{\p[\theta]{z_t}{z\tm,s_t,u\tm}}}}.
\end{aligned}
\raisetag{43pt}
\end{gather} 
The full derivation can be found in appendix~\ref{app:lower-bound}. 
We generally approximate the expectations with one sample by using the reparametrization trick, the exception being the KL between two Concrete random variables in which case we take 10 samples for the approximation.
For the KL on the switching variables, we further introduce a scaling factor $\beta < 1$ (as first suggested in \citet{Higgins2016}, although they suggested increasing the weighting of the KL term) to scale down its importance.
This decision can be justified by other work which has demonstrated theoretically and empirically the inadequacy of maximum likelihood training on the ELBO for learning latent representations \citep{alemi2018fixing}.
More details on the training procedure can be found in appendix~\ref{app:training}.

\section{Experiments}

In this section, we evaluate our approach on a diverse set of physics simulations based on partially observable system states or high-dimensional images as observations. We show that our model outperforms previous models and that our switching variables learn meaningful representations. 

Models we compare to are Deep Variational Bayes Filter (DVBF) \citep{karl2017deep}, DVBF Fusion \citep{karl2017unsupervised} (called fusion as they do the same Gaussian multiplication in the inference network) which is closest to our model but does not have a stochastic treatment of the transition, the Kalman VAE (KVAE) \citep{fraccaro2017disentangled} and a vanilla LSTM \citep{hochreiter1997long}.

\begin{table*}[t]
\caption{Mean squared error (MSE) on predicting future observations. Static refers to constantly predicting the first observation of the sequence.}
\label{table:mse-roboschool}
\begin{center}
\begin{small}
\begin{sc}
\begin{tabular}{l|ccc|ccc}
\toprule
 & \multicolumn{3}{|c}{\textbf{Reacher}} & \multicolumn{3}{|c}{\textbf{3-Ball Maze}}\\
\midrule
Prediction steps & 1 & 5 & 10 & 1 & 5 & 10 \\
\midrule
Static & 5.80E-02 & 5.36E-01 & 1.25E+00 & 1.40E-02 & 5.74E-01 & 2.65E+00 \\
LSTM & 3.07E-02 & 3.67E-01 & 1.02E+00 & 7.20E-03 & 1.58E-01 & 2.60E-01 \\
DVBF & 1.10E-02 & 3.06E-01 & 6.05E-01 & 6.20E-03 & 1.36E-01 & 1.82E-01 \\
DVBF Fusion & 4.90E-03 & 2.97E-02 & 8.25E-02 & 4.33E-03 & 2.03E-02 & 4.88E-02 \\
Ours (Concrete) & 1.06E-02 & 5.73E-02 & 1.56E-01 & 2.28E-03 & 1.22E-02 & 3.40E-02 \\
Ours (Normal) & \textbf{3.39E-03} & \textbf{1.85E-02} & \textbf{4.97E-02} & \textbf{1.30E-03} & \textbf{5.52E-03} & \textbf{1.38E-02} \\
\bottomrule
\end{tabular}
\end{sc}
\end{small}
\end{center}
\end{table*}

\begin{figure}[h]
  \centering 
  \includegraphics[scale=.175]{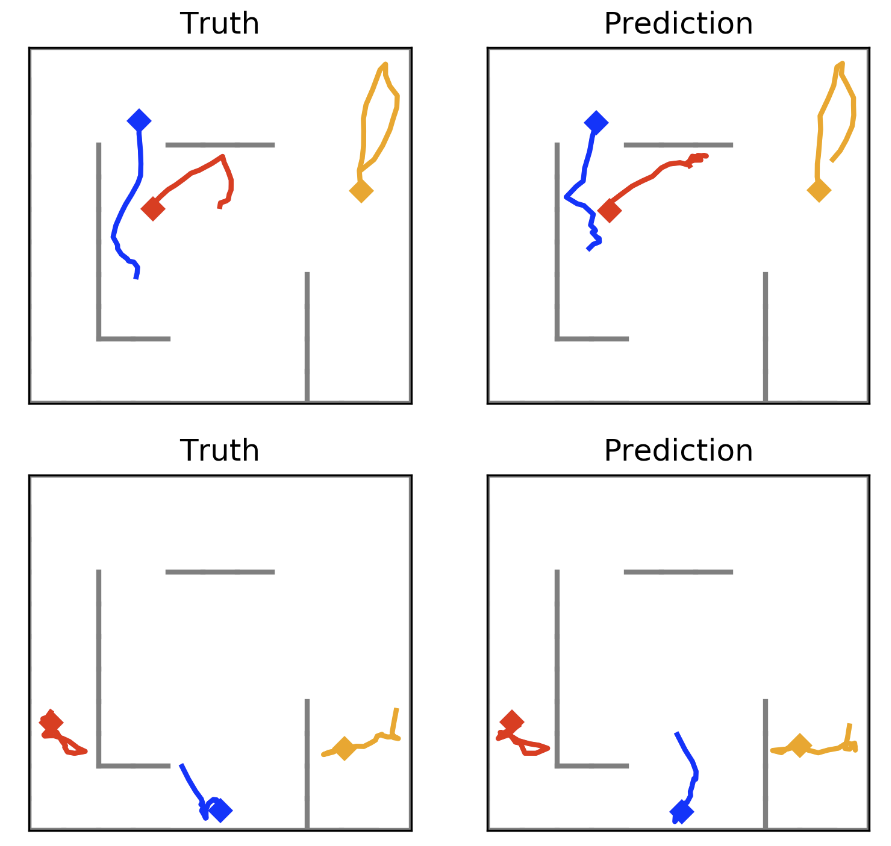}
  \label{fig:multi_agent_maze_pred_trajectory}
  \caption{Comparison of actual and predicted 20 step trajectories. The diamond marker denotes the starting position of a trajectory. These results have been produced with Concrete switching variables.}
\end{figure}
\subsection{Multiple Bouncing Balls in a Maze}\label{sec:bouncing-ball}
Our first experiment is a custom 3-agent maze environment simulated with Box2D. 
Each agent is fully described by its $(x,y)$-coordinates and its current velocity and may accelerate in either direction.
We learn in a partially observable setting and limit the observations to the agents' positions, therefore $x \in  \RR^{6}$ while the true state space is in $\RR^{12}$ and $u \in \RR^{6}$. 

Our first objective is to evaluate the learned latent space.
We start by training a linear regression model on the latent space $z$ to see if we have recovered a linear encoding of the unobserved velocities. 
Here, we achieve an R2 score of 0.92 averaged over all agents and velocity directions.

We now shift our focus to the switching variables that we anticipated to encode interactions with walls.
We provide a qualitative confirmation of that in figure~\ref{fig:maze} where we see switching variables encoding space where there is no interaction in the next time step and variables which encode walls, distinguishing between vertical and horizontal ones.
In figure~\ref{fig:maze-deterministic} one can see that if the choice of locally linear transition is treated deterministically, we do not learn global features of the same kind. 
To confirm our visual inspection, we train a simple decision tree based on latent space $s$ in order to predict an interaction with a wall. 
Here, we achieve an F1 score of $0.46$. 
It is difficult to say what a good value should look like as collisions with low velocity are virtually indistinguishable from no collision, but certainly a significant portion of collisions has been successfully captured.
We were unable to capture collisions between two agents which can be explained by their rare occurrence and much more complicated ensuing dynamics.

Going over to quantitative evaluation, we compare our prediction quality to several other methods in table~\ref{table:mse-roboschool} where we outperform all of our chosen baselines.
Also, curiously, modelling switching variables by a Gaussian distribution outperforms the Concrete distribution in all of our experiments.
Aside from known practical issues with training a discrete variable via backpropagation, we explore one reason why that may be in section~\ref{sec:delta_t}, which is the greater susceptibility to the chosen scale of temporal discretization.

\subsection{Reacher}

\begin{figure}[h]
  \centering 
  \begin{subfigure}[b]{0.24\textwidth}
  \centering
  	\includegraphics[scale=0.33]{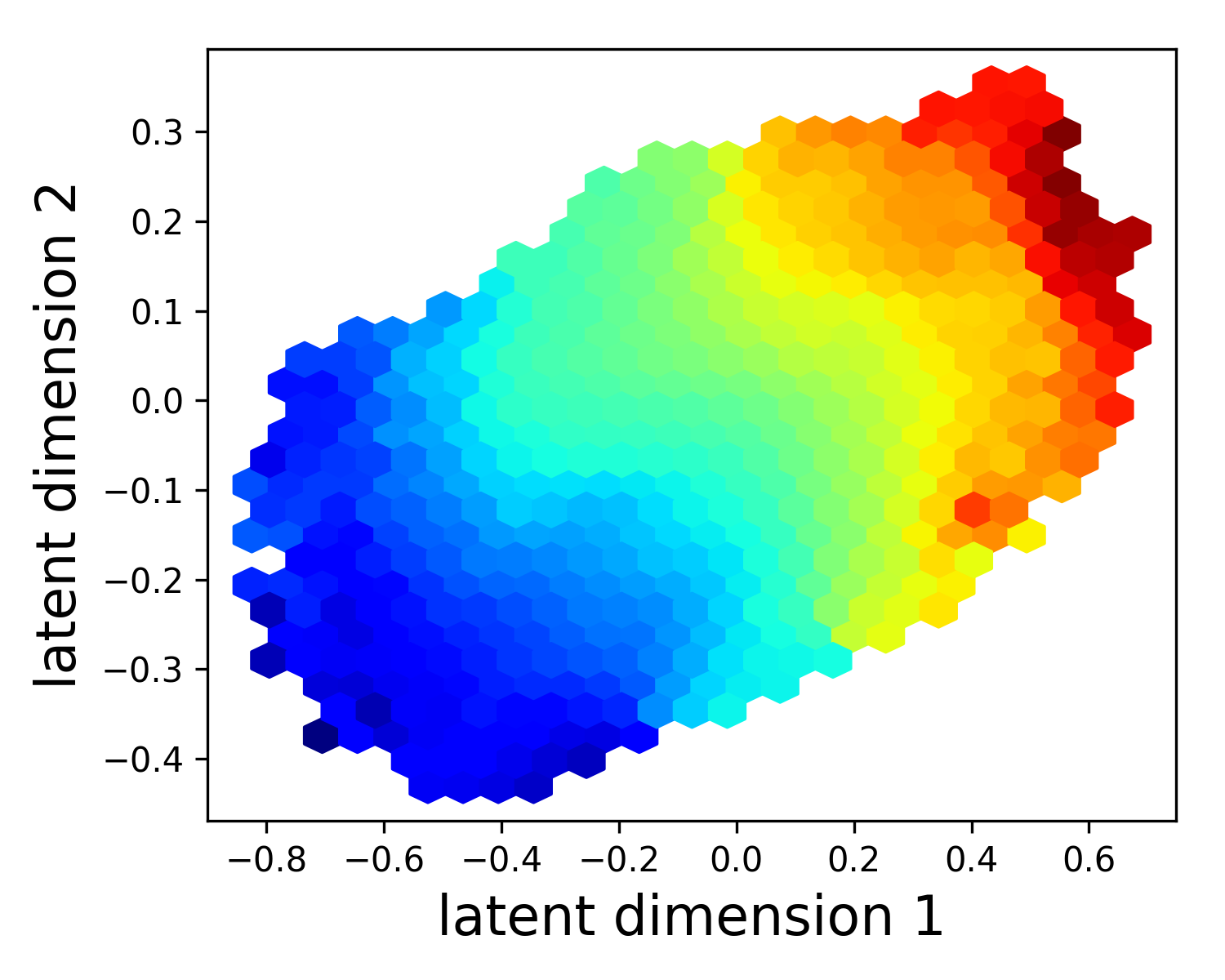}
    \caption{Encoding shoulder velocity.}
    \label{fig:reacher-z-space}
  \end{subfigure}\hfill
  \begin{subfigure}[b]{0.24\textwidth}
  \centering
  	\includegraphics[scale=0.33]{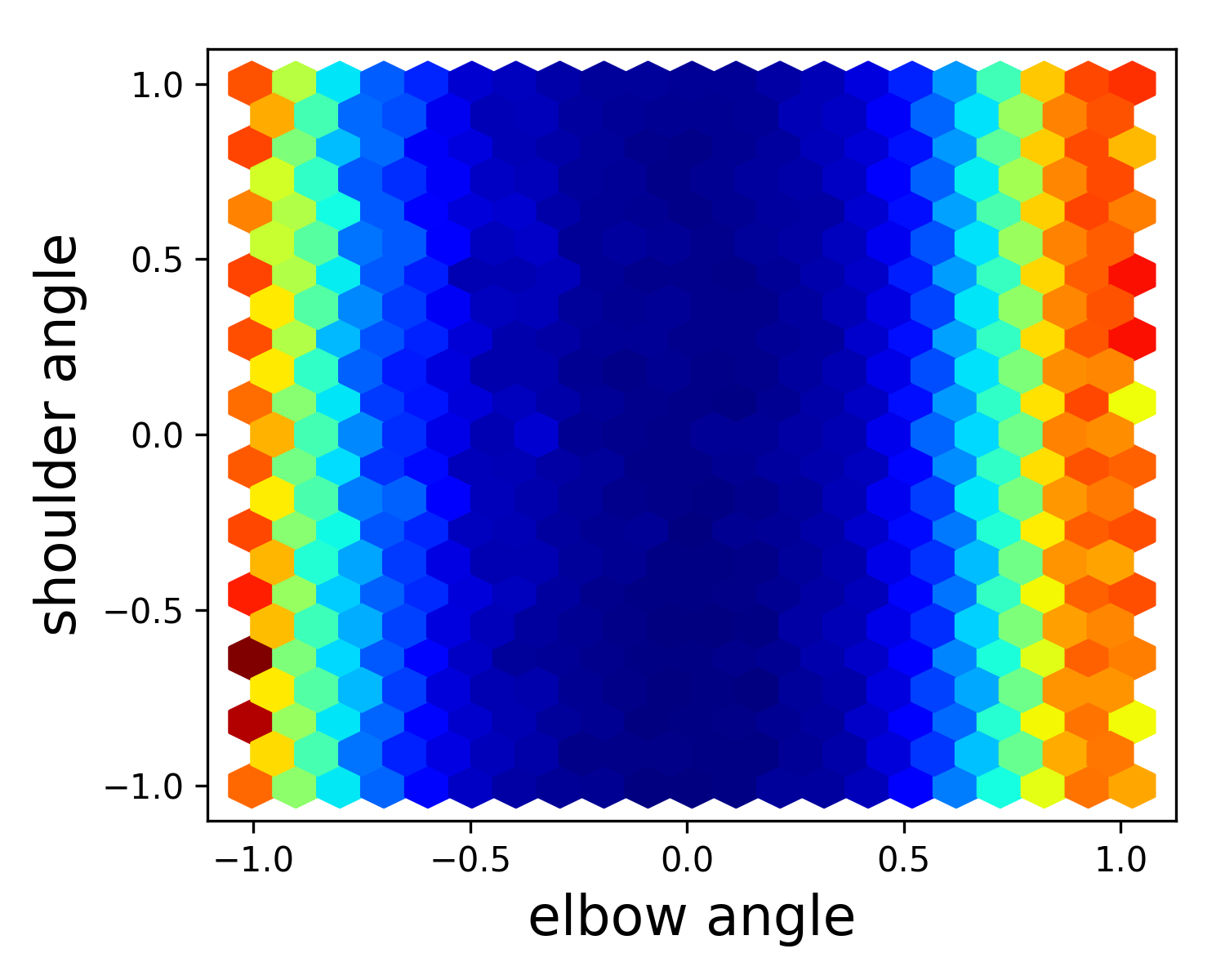}
    \caption{Encoding joint collisions.}
    \label{fig:reacher-s-space}
  \end{subfigure}
  \caption{Encoding features of the reacher environment. Figure (a) shows two latent dimensions colored by the ground truth shoulder velocity. The model captures the shoulder's velocity purely out of provided joint angle data. Figure (b) highlights the activation of a Concrete switching variable. Note that the elbow's angle is provided relative to the upper link, meaning that a (normalized) value of -1 or 1 leads to a collision with the upper link.}
  \label{fig:reacher}
\end{figure}

We then evaluate our model on the RoboschoolReacher \citep{roboschool} environment. 
Again, we learn only on partial observations, removing velocities and leaving us with just the positions or angles of the joints as observations.
The reacher's dynamics are globally the same unless the upper and lower joints collide which is again a feature we expect our switching variables to detect.
Similar to before, we inspect the learned latent space in figure~\ref{fig:reacher} visually where  show linear encoding of shoulder joint's velocity and encoding of collision dynamics.
The chosen dynamics are unaffected by the shoulder angle, but are sensitive to the relative angle of the elbow to the upper link.
To confirm our qualitative analysis, we again learn a linear classifier based on latent space $s$ and reach an F1 score of $0.53$.
The predictive quality of our model is compared to other methods in table~\ref{table:mse-roboschool}.

\subsection{Ball in a Box on Image Data}
\begin{figure*}[t]
\resizebox{\textwidth}{!}{
\includegraphics[scale=0.5]{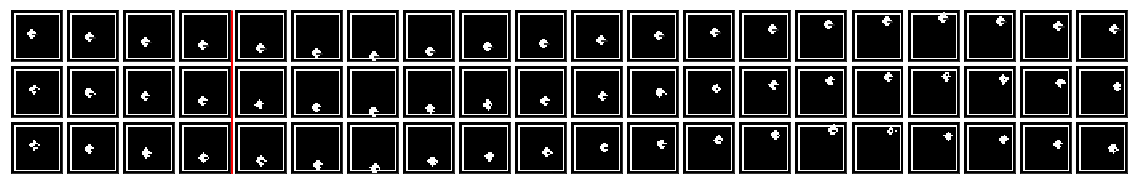}}
\caption{First row: data, second row: filtered reconstructions, third row: predictions. The first 4 steps are used to find a stable starting state, predictions start with step 5 (after the red line).}
\label{fig:kvae-box-sample}
\end{figure*}

Here, we evaluate our method on high-dimensional image observations using the single bouncing ball environment used by \citet{fraccaro2017disentangled}. 
They simulated 5000 sequences of 20 time steps each of a ball moving in a two-dimensional box, where each video frame is a $32 \times 32$ binary image. 
There are no forces applied to the ball, except for the fully elastic collisions with the walls. 
Initial position and velocity are randomly sampled.

\begin{figure*}[t]
  \centering 
  \begin{subfigure}[b]{0.32\textwidth}
  \centering
  \resizebox{!}{.73\textwidth}{
   \begin{tikzpicture}[scale=1]
	\begin{axis}[xlabel=steps predicted into the future, ylabel=Fraction of incorrect pixels,legend style={at={(0.03,.85)},anchor=west},ymin=0, every axis plot/.append style={very thick}] 
	\addplot[color=orange,mark=square,densely dotted] coordinates { 
		(1, 0.003) (5, 0.009) (10, 0.015) (15, 0.023)
	}; 
	\addlegendentry{KVAE (smoothed)}
	\addplot[color=darkgray,mark=x,dashed] coordinates { 
		(1, 0.01) (5, 0.015) (10, 0.02) (15, 0.027)
	}; 
	\addlegendentry{KVAE (generative)}
	\addplot[color=blue,mark=diamond] coordinates { 
		(1, 0.011) (5, 0.014) (10, 0.017) (15, 0.021)
	}; 
	\addlegendentry{Ours (generative)}
	\end{axis} 
	\end{tikzpicture}}
	\caption{Fraction of incorrectly predicted pixels.}
    \label{fig:kave-box-pixel-fraction}
  \end{subfigure}\hfill
  \begin{subfigure}[b]{0.32\textwidth}
   \centering
    \resizebox{!}{.7\textwidth}{
   \begin{tikzpicture}[scale=1]
	\begin{axis}[xlabel=prediction horizon, ylabel=$R^2$,legend style={at={(0.03,.15)},anchor=west},ymin=0.7,ymax=1, every axis plot/.append style={very thick}] 	
	\addplot[color=orange,mark=square,densely dotted] coordinates { 
		(1,0.999) (2, 0.995) (5, 0.99) (10, 0.985) (15, 0.98) (20, 0.97) (25, 0.95) (30, 0.94)
	}; 
	\addlegendentry{TrSLDS}
	
	\addplot[color=darkgray,mark=x,dashed] coordinates { 
		(1, 0.999) (2, 0.99) (5, 0.98) (10, 0.98) (15, 0.965) (20, 0.94) (25, 0.84) (30, 0.76)
	}; 
	\addlegendentry{rSLDS}
	
	\addplot[color=blue,mark=diamond] coordinates {
		(1, 0.999) (2, 0.999) (5, 0.999) (10, 0.996) (15, 0.992) (20, 0.981) (25, 0.969) (30, 0.956)
	}; 
	\addlegendentry{Ours (Concrete)}
	\end{axis} 
	\end{tikzpicture}}
    \caption{FitzHugh-Nagumo multi-step prediction.}
    \label{fig:fhn-comparison}
  \end{subfigure}\hfill
  \begin{subfigure}[b]{0.32\textwidth}
  \centering
    \resizebox{!}{.7\textwidth}{
   \begin{tikzpicture}[scale=1]
	\begin{axis}[xlabel=$\Delta$ t (log scale), ylabel=Absolute Error (log scale),legend style={at={(0.03,.85)},anchor=west}, xmode=log, ymode=log, every axis plot/.append style={very thick}] 
	\addplot[color=darkgray,mark=x,dashed] coordinates { 
		(0.002, 1.10E-02) (0.005, 1.84E-02) (0.01, 2.78E-02) (0.02, 0.0614) (0.04, 0.1531) (0.1, 0.205)
	}; 
	\addlegendentry{Normal}
	\addplot[color=blue,mark=diamond] coordinates { 
		(0.002, 6.46E-03) (0.005, 1.32E-02) (0.01, 4.60E-02) (0.02, 0.1012) (0.04, 0.2857) (0.1, 0.3318)
	}; 
	\addlegendentry{Concrete}
	\end{axis} 
	\end{tikzpicture}}
    \caption{Susceptibility to discretization scale.}
    \label{fig:delta-t-scale}
  \end{subfigure}
  \caption{(a) Our dynamics model is outperforming even the smoothed KVAE for longer trajectories. (b) Our models performs as well as TrSLDS. (c) Modelling switching variables as Concrete random variables scales less favourably with increasing time discretization intervals.}
\end{figure*}
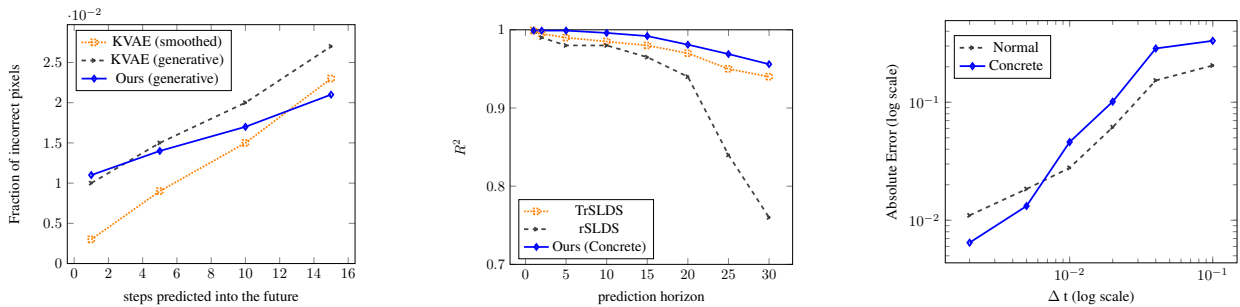

In figure~\ref{fig:kave-box-pixel-fraction} we compare our model to both the smoothed and generative version of the KVAE. 
The smoothed version receives the final state of the trajectory after the $n$ predicted steps which is fed into the smoothing capability of the KVAE.
One can see that our model learns a better transition model, even outperforming the smoothed KVAE for longer sequences.
For short sequences, KVAE performs better which highlights the value of it disentangling the latent space into separate object and dynamics representation.
A sample trajectory is plotted in figure~\ref{fig:kvae-box-sample}.

\subsection{FitzHugh-Nagumo}
To compare to recurrent SLDS (rSLDS, \citet{Linderman2016}) and tree-structured SLDS (TrSLDS, \citet{nassar2018tree}), we adopt their FitzHugh-Nagumo (FHN, \citet{fitzhugh1961impulses}) experimental setup.
FHN is a 2-dimensional relaxation oscillator commonly used throughout neuroscience to describes a prototype of an excitable system (e.g.,\ a neuron). It is fully described by the following system of differential equations:
\begin{equation}
\begin{gathered}
\dot{v} = v - \frac{v^3}{3} - w + I_{\mathrm{ext}}, \quad \tau \dot{w} = v + a - bw
\end{gathered}
\end{equation}
Following their setup, we set the parameters to $a=0.7$, $b=0.8$, $\tau=12.5$ and external stimulus $I_{\mathrm{ext}} \sim \gauss{0.7, 0.04}$. 
We create 100 trajectories of length 430 where the last 30 time steps are withheld during training and used for evaluation. 
Starting states are drawn uniformly from $[-3;3]^2$.
In figure~\ref{fig:fhn-comparison}, we compare the models based on normalized multi-step predictive performance where our model matches the performance of TrSLDS.

\subsection{Susceptibility to the Scale of Temporal Discretization}\label{sec:delta_t}
In this section, we would like to explore how the choice of $\Delta t$ when discretizing a system influences our results. 
This is a crucial factor often ignored or presumed to be chosen appropriately, although there has been some recent work addressing this issue specifically \citep{jayaraman2018time, neitz2018adaptive}.
 
In particular, we suspect our model with discrete (Concrete) switching latent variables to be more susceptible to scaling than when modelled by a normal distribution. 
This is because in the latter case the switching variables can scale the various matrices more freely, while in the former scaling up one system necessitates scaling down another.
For empirical comparison, we go back to our custom maze environment (this time with only one agent as this is not pertinent to our question at hand) and learn the dynamics on various discretization scales. 
Then we compare the absolute error's growth for both approaches in figure~\ref{fig:delta-t-scale} which supports our hypothesis. 
While the discrete approximation even outperforms for small $\Delta t$, there is a point where it rapidly becomes worse and gets overtaken by the normally distributed approximation.
This suggests that $\Delta t$ was simply chosen to be too large in both the reacher and the ball in a box with image observations experiment.

\section{Conclusion}

We have shown that our construction of using switching variables encourages learning a richer and more interpretable latent space. 
In turn, the richer representation led to an improvement of simulation accuracy in various tasks.
In the future, we would like to look at other ways to approximate the discrete switching variables and exploit this approach for model-based control on real hardware systems.
Furthermore, addressing the open problem of disentangling latent spaces is essential to fitting simple dynamics and would lead to significant improvements of this approach.

\FloatBarrier
\bibliography{bibliography}
\bibliographystyle{icml2019}
\clearpage

\appendix
\begin{table*}[t]
\caption{Dimensionality of environments.}
\label{tab:environments}
\begin{center}
\begin{tabular}{lccc}
Dimensionality of & Observation Space & Control Input Space & Ground Truth State Space  \\
\midrule
Reacher 			& $7$ & $2$ & $9$ \\
Hopper	 			& $8$ & $3$ & $15$ \\
Multi Agent Maze 	& $6$ & $6$ & $12$ \\
Image Ball in Box	& $32 \times 32$ & $0$ & $4$ \\
FitzHugh-Nagumo		& $2$ & $1$ & $2$\\
\end{tabular}
\end{center}
\end{table*}
\section{Lower Bound Derivation}\label{app:lower-bound}
For brevity we omit conditioning on control inputs $u\Ts$.
\begin{equation*}
\allowdisplaybreaks
\begin{aligned}
\log \: &\p{x\Ts} = \log \int_{z\Ts} \int_{s_{2:T}} \q[\phi]{z\Ts, s_{2:T}}{x\Ts}\\
&\qquad\frac{\p[\theta]{x\Ts}{z\Ts} \p[\theta]{z\Ts,s_{2:T}} }{\q[\phi]{z\Ts, s_{2:T}}{x\Ts}}\\
\geq& \int_{z\Ts} \int_{s_{2:T}} \q[\phi]{z\Ts, s_{2:T}}{x\Ts}\\
&\qquad\log \frac{\p[\theta]{x\Ts}{z\Ts} \p[\theta]{z\Ts,s_{2:T}} }{\q[\phi]{z\Ts, s_{2:T}}{x\Ts}}\\
=& \int_{z\Ts} \int_{s_{2:T}} \q[\phi]{z\Ts, s_{2:T}}{x\Ts} \log \p[\theta]{x_t}{z_t,s_t}\\
&+\int_{z\Ts} \int_{s_{2:T}} \q[\phi]{z\Ts, s_{2:T}}{x\Ts}\\
&\qquad\qquad\log \frac{\p[\theta]{z\Ts,s_{2:T}} }{\q[\phi]{z\Ts, s_{2:T}}{x\Ts}}\\
=& \int_{z_1} \int_{s_2} \dots \int_{s_T} \int_{z_T} \q{z_1}{x\Ts} \q{s_2}{z_1, x\Ts} \dots \\
&\dots\q{s_T}{z\Tm, s\Tm, x\Ts} \q{z_T}{z\Tm, s_T, x\Ts}\\ 
& \log \p[\theta]{x_1}{z_t,s_t} \dots \p[\theta]{x_T}{z_T,s_T} \\
&- \kl{\q{z\Ts,s_{2:T}}{x\Ts}}{\p[\theta]{z\Ts,s_{2:T}}}\\
=& \expc[z_1 \sim \q{z_1}{\cdot}]{\p{x_1}{z_1}} \\
&+ \sum_{t=2}^T \expc[s_t \sim \q{s_t}{\cdot}]{\expc[z_t \sim \q{z_t}{\cdot}]{\p{x_t}{z_t,s_t}}}\\ 
&- \kl{\q{z\Ts,s_{2:T}}{x\Ts}}{\p{z\Ts,s_{2:T}}}
\end{aligned}
\end{equation*}

\subsection{Factorization of the KL Divergence}
The dependencies on data $x\Ts$ and $u\Ts$ as well as parameters $\phi$ and $\theta$ are omitted in the following for convenience.
\begin{equation*}
\begin{aligned}
&\kl{\q{z_1,s_2,\dots,s_T,z_T}}{\p{z_1,s_2,\dots,s_T,z_T}} \\
&\textit{(Factorization of the variational approximation)}\\ 
=& \int_{z_1} \int_{s_2} \dots \int_{s_T} \int_{z_T} \q{z_1} \q{s_2}{z_1} \dots \\
&\qquad \dots\q{s_T}{z\Tm, s\Tm} \q{z_T}{z\Tm, s_T}\\ 
&\log \frac{\q{z_1} \q{s_2}{z_1} \dots \q{z_T}{z\Tm, s_T}}{\p{z_1,s_2,\dots,s_T,z_T}}\\
&\textit{(Factorization of the prior)}\\ 
=& \int_{z_1} \int_{s_2} \dots \int_{s_T} \int_{z_T} \q{z_1} \q{s_2}{z_1} \dots \\
&\qquad\dots\q{s_T}{z\Tm, s\Tm} \q{z_T}{z\Tm, s_T}\\
&\log \frac{\q{z_1} \q{s_2}{z_1} \dots \q{z_T}{z\Tm, s_T}}{\p{z_1}\p{s_2}{z_1} \dots \p{z_T}{z\Tm,s_T}}\\
\end{aligned}
\end{equation*}
\begin{equation*}
\begin{aligned}
&\textit{(Expanding the logarithm by the product rule)}\\
=& \int_{z_1} \q{z_1} \log \frac{\q{z_1}}{\p{z_1}}\\
&+ \int_{z_1}\int_{s_2} \q{z_1}\q{s_2}{z_1} \log \frac{\q{s_2}{z_1}}{\p{s_2}{z_1}} \\
&+ \sum_{t=2}^T \int_{z_1} \int_{s_2} \dots \int_{s_T} \int_{z_T} \q{z_1} \q{s_2}{z_1} \dots\\
&\qquad\dots\q{z_T}{z\Tm, s_T} \log \frac{\q{z_t}{z\tm, s_t}}{\p{z_t}{z\tm, s_t}}\\
&+ \sum_{t=3}^T \int_{z_1} \int_{s_2} \dots \int_{s_T} \int_{z_T} \q{z_1} \q{s_2}{z_1} \dots\\
&\qquad\dots\q{z_T}{z\Tm, s_T} \log \frac{\q{s_t}{z\tm, s\tm}}{\p{s_t}{z\tm, s\tm}}\\
&\textit{(Ignoring constants)}\\
=& \kl{\q{z_1}}{\p{z_1}}\\
&+ \expc[z_1 \sim \q{z_1}]{\kl{\q{s_2}{z_1}}{\p{s_2}{z_1}}} \\
&+ \sum_{t=2}^{T-1} \EE_{z\tm \sim q(z\tm |\cdot)} \Big[ \EE_{s_t \sim q(s_t |\cdot)} \big[\\
&\qquad\quad {\kl{\q{z_t}{z\tm, s_t}}{\p{z_t}{z\tm, s_t}}} \big]\Big]\\
&+ \sum_{t=3}^{T-1} \EE_{s\tm \sim q(s\tm |\cdot)} \Big[ \EE_{z\tm \sim q(z\tm |\cdot)} \big[ \\
&\qquad\quad \kl{\q{s_t}{z\tm, s\tm}}{\p{s_t}{z\tm, s\tm}} \big]\Big]\\
\end{aligned}
\end{equation*}
\section{Comparison to Previous Models}\label{sec:discussion}
We want to emphasize some subtle differences to previously proposed architectures that make an empirical difference, in particular for the case when $s_t$ is chosen to be continuous.
In \citet{watter2015embed} and \citet{karl2017deep}, the latent space is already used to draw transition matrices, however they do not extract features such as walls or joint constraints.
There are a few key differences to our approach.

First, our latent switching variables $s_t$ are only involved in predicting the current observation $x_t$ through the transition selection process.
The likelihood model therefore does not need to learn to ignore some input dimensions which are only helpful for reconstructing future observations but not the current one.

There is also a clearer restriction on how $s_t$ and $z_t$ may interact: $s_t$ may now only influence $z_t$ by determining the dynamics, while previously $z_t$ influenced both the choice of transition function as well as acted inside the transition itself.
These two opposing roles lead to conflicting gradients as to what should be improved.
Furthermore, the learning signal for $s_t$ is rather weak so that scaling down the KL-regularization was necessary to detect good features.

Lastly, a locally linear transition may not be a good fit for variables determining dynamics as such variables may change very abruptly.
Therefore, it might be beneficial to have part of the latent space evolve according to locally linear dynamics and other parts according to a general purpose neural network transition.
Overall, our structure of choosing a transition gives a stronger bias towards learning such features when compared to other methods.
\section{Training}\label{app:training}
Overall, training the Concrete distribution has given us the biggest challenge as it was very susceptible to various hyperparameters. 
We made use of the fact that we can use a different temperature for the prior and approximate posterior \citep{Maddison2016} and we do independent hyperparameter search over both. 
For us, the best values were $0.75$ for the posterior and $2$ for the prior. 
Additionally, we employ an exponential annealing scheme for the temperature hyperparameter of the Concrete distribution. 
This leads to a more uniform combination of base matrices early in training which has two desirable effects. 
First, all matrices are scaled to a similar magnitude, making initialization less critical. 
Second, the model initially tries to fit a globally linear model, leading to a good starting state for optimization.\\
With regards to optimizing the KL-divergence, there is no closed-form analytical solution for two Concrete distributions. 
We therefore had to resort to a Monte Carlo estimation with $n$ samples where we tried $n$ between $1$ and $1000$. While using a single samples was (numerically) unstable, using a large number of samples also didn't result in observable performance improvements.
We therefore settled on using $10$ samples for all experiments.

In all experiments, we train everything end-to-end with the ADAM optimizer \citep{Kingma2014Adam}.
We start with learning rate of $5\mathrm{e}{-4}$ and use an exponential decay schedule with rate $\lambda \in \lbrace 0.95, 0.97, 0.98 \rbrace$ (see table~\ref{tab:hyperparameters}) every $2000$ iterations. 
\section{Experimental Setup}\label{app:experiments}

\subsection{Data Set Generation}\label{app:hyperparameters}
\begin{table*}[t]
\caption{Overview of hyperparameters.}
\label{tab:hyperparameters}
\begin{center}
\begin{tabular}{lccccc}
 & Multi Agent Maze & Reacher & Image Ball in Box & FitzHugh-Nagumo \\
\midrule
\# episodes & 50000 & 20000 & 5000 & 100\\
episode length & 20 & 30 & 20 & 400 \\
batch size & 256 & 128 & 256 & 32\\
dimension of $z$ & 32 & 16 & 8 & 4\\
dimension of $s$ & 16 & 8 & 8 & 8\\
posterior temperature & 0.75 & 0.75 & 0.67 & 0.67\\
prior temperature & 2 & 2 & 2 & 2\\
temperature annealing steps & 100 & 100 & 100 & 100\\
temperature annealing rate & 0.97 & 0.97 & 0.98 & 0.95\\
$\beta$ (KL-scaling of switching variables) & 0.1 & 0.1 & 0.1 & 0.1  \\
\end{tabular}
\end{center}
\end{table*}

\subsubsection{Roboschool Reacher}
To generate data, we follow a Uniform distribution $\mathcal{U} \sim [-1,1]$ as the exploration policy. Before we record data, we take 20 warm-up steps in the environment to randomize our starting state. We take the data as is without any other preprocessing.

\subsubsection{Multi Agent Maze}
Observations are normalized to be in $[-1,1]$. Both position and velocity is randomized for the starting state. We again follow a Uniform distribution $\mathcal{U} \sim [-1,1]$ as the exploration policy.

\subsection{Network Architecture}\label{app:network-architecture}

For most networks, we use MLPs implemented as residual nets \citep{he2016deep} with ReLU activations. 

Networks used for the reacher and maze experiments.
\begin{itemize}
\item $\q[\mathrm{meas}]{z_t}{x_{\geq t},u_{\geq t}}$: MLP consisting of two residual blocks with 256 neurons each. We only condition on the current observation $x_t$ although we could condition on the entire sequence. This decision was taken based on empirical results.
\item $\q[\mathrm{trans}]{z_t}{z\tm, u\tm, s_t}$: In the case of Concrete random variables, we just combine the base matrices and apply the transition dynamics to $z\tm$. For the Normal case, the combination of matrices is preceded by a linear combination with softmax activation. (see equation~\ref{eq:softmax-alpha})
\item $\q[\mathrm{meas}]{s_t}{x_{\geq t},u_{\geq t}}$: is implemented by a backward LSTM with 256 hidden units. We reuse the preprocessing of $\q[\mathrm{meas}]{z_t}{x_t}$ and take the last hidden layer of that network as the input to the LSTM.
\item $\q[\mathrm{trans}]{s_t}{s\tm, z\tm, u\tm}$: MLP consisting of one residual block with 256 neurons.
\item $\q[\mathrm{initial}]{w}{x\Ts,u\Ts}$: MLP consisting of two residual block with 256 neurons optionally followed by a backward LSTM. We only condition on the first 3 or 4 observations for our experiments.
\item $\q[\mathrm{initial}]{s_2}{x\Ts,u\Ts}$: The first switching variable in the sequence has no predecessor. We therefore require a replacement for $\q[\mathrm{trans}]{s_t}{s\tm, z\tm, u\tm}$ in the first time step, which we achieve by independently parameterizing another MLP.
\item $\p{x_t}{z_t}$: MLP consisting of two residual block with 256 neurons.
\item $\p{z_t}{z\tm, u\tm, s_t}$: Shared parameters with $\q[\mathrm{trans}]{z_t}{z\tm, u\tm, s_t}$.
\item $\p{s_t}{s\tm, z\tm, u\tm}$: Shared parameters with $\q[\mathrm{trans}]{s_t}{s\tm, z\tm, u\tm}$.
\end{itemize}

We use the same architecture for the image ball in a box experiment, however we increase number of neurons of $\q[\mathrm{meas}]{z_t}{x_{\geq t},u_{\geq t}}$ to 1024.

For the FitzHugh-Nagumo model we downsize our model and restrict all networks to a single hidden layer with 128 neurons.

\section{On Scaling Issues of Switching Linear Dynamical Systems}\label{app:scaling}
Let's consider a simple representation of a ball in a rectangular box where its state is represented by its position and velocity. 
Given a small enough $\Delta t$, we can approximate the  dynamics decently by just $3$ systems: no interaction with the wall, interaction with a vertical or horizontal wall (ignoring the corner case of interacting with two walls at the same time).
Now consider the growth of required base systems if we increase the number of balls in the box (even if these balls cannot interact with each other). 
We would require a system for all combinations of a single ball's possible states: $3^2$. 
This will grow exponentially with the number of balls in the environment.

One way to alleviate this problem that requires only a linear growth in base systems is to independently turn individual systems on and off and let the resulting system be the sum of all activated systems. 
A base system may then represent solely the transition for a single ball being in specific state, while the complete system is then a combination of $N$ such systems where $N$ is the number of balls.
Practically, this can be achieved by replacing the \textit{softmax} by a \textit{sigmoid} activation function or by replacing the categorical variable $s$ of dimension $M$ by $M$ Bernoulli variables indicating whether a single system is active or not. 
We make use of this parameterization in the multiple bouncing balls in a maze environment.

Theoretically, a preferred approach would be to disentangle multiple systems (like balls, joints) and apply transitions only to their respective states. 
This, however, would require a proper and unsupervised separation of (mostly) independent components. We defer this to future work.

\section{Further Results}\label{app:results}

\begin{figure*}[b]
  \centering 
  \begin{subfigure}[b]{\textwidth}
  \centering
  \resizebox{\textwidth}{!}{
   \includegraphics[scale=1.]{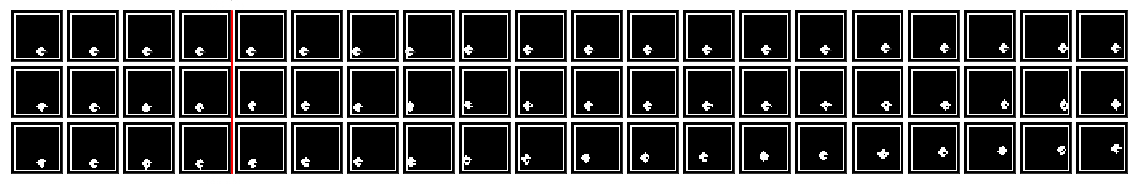}
   }
  \end{subfigure}
  \begin{subfigure}[b]{\textwidth}
  \centering
    \resizebox{\textwidth}{!}{
   \includegraphics[scale=1.]{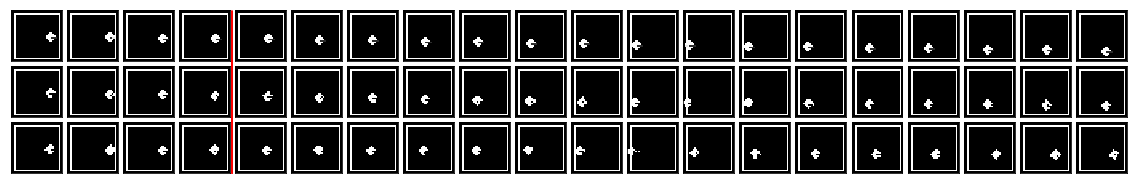}
   }
  \end{subfigure}
  \begin{subfigure}[b]{\textwidth}
  \centering
    \resizebox{\textwidth}{!}{
   \includegraphics[scale=1.]{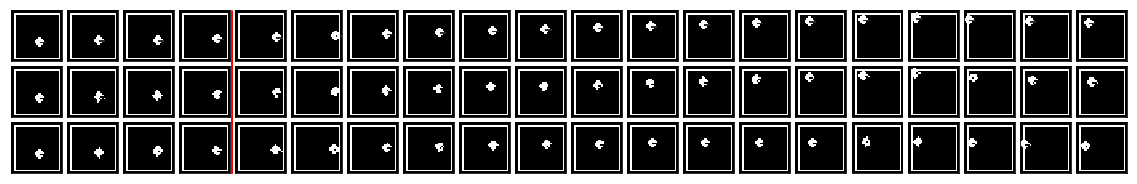}
   }
  \end{subfigure}
  \begin{subfigure}[b]{\textwidth}
  \centering
    \resizebox{\textwidth}{!}{
   \includegraphics[scale=1.]{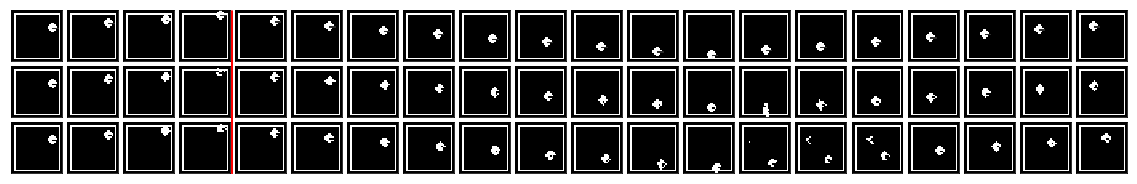}
   }
  \end{subfigure}
  \caption{First row: data, second row: reconstructions, third row: predictions. The first 4 steps are used to find a stable starting state, predictions start with step 5 (after the red line). These results have been produced with Gaussian distributed switching variables.}
\end{figure*}

\end{document}